\documentclass{article}

\usepackage[final]{neurips_2019}

\usepackage[utf8]{inputenc} % allow utf-8 input
\usepackage[T1]{fontenc}    % use 8-bit T1 fonts
\usepackage{hyperref}       % hyperlinks
\usepackage{url}            % simple URL typesetting
\usepackage{booktabs}       % professional-quality tables
\usepackage{amsfonts}       % blackboard math symbols
\usepackage{nicefrac}       % compact symbols for 1/2, etc.
\usepackage{microtype}      % microtypography
\usepackage{times}
\usepackage{latexsym}
\usepackage{subcaption,booktabs}
\usepackage{graphicx}
\usepackage{amsmath}
\usepackage{url}
\usepackage{float}

\title{On Compositionality in Neural Machine Translation}

\author{
  Vikas Raunak\thanks{Equal Contribution}\\
  \texttt{vraunak@andrew.cmu.edu}
  \And
  Vaibhav Kumar$^{*}$\\
  \texttt{vaibhav2@andrew.cmu.edu}\\
  \And
   Florian Metze\\
  \texttt{fmetze@andrew.cmu.edu}
}

\begin{document}

\maketitle

\begin{abstract}
    We investigate two specific manifestations of compositionality in Neural Machine Translation (NMT) : (1) Productivity - the ability of the model to extend its predictions beyond the observed length in training data and (2) Systematicity - the ability of the model to systematically recombine known parts and rules. We evaluate a standard Sequence to Sequence model on tests designed to assess these two properties in NMT. We quantitatively demonstrate that inadequate temporal processing, in the form of poor encoder representations is a bottleneck for both Productivity and Systematicity. Motivated by the anslysis, we propose a simple pre-training mechanism which leads to a significant improvement in BLEU scores.
\end{abstract}

\vspace{-0.7cm}
\section{Introduction}
\vspace{-0.3cm}
Sequence-to-sequence (Seq2Seq) networks have achieved impressive results \citep{bahdanau2014neural, sutskever2014sequence, neubig2017neural,vinyals2015neural,venugopalan2015sequence,karpathy2015deep} on a variety of problems within natural language processing. However, their inability to handle long sentences \citep{length1} as well as a lack of systematic generalizability \citep{lake2017still} questions the ability of seq2seq networks to model compositionality in natural language. \citet{hupkes2019compositionality} provide tests corresponding to different ineterpretations of compositionality. In this work, we investigate two such properties, which are intrinsic to the way that humans utilize language namely, - (1) Productivity: the ability to generalize beyond the observed length and (2) Systematicity: the ability to recombine knows parts and rules. Loosely, Productivity could be regarded as "unbounded" application of known rules, while systematicity implies arbitrary recombinations of known rules.

In the coming sections, we first describe the quantitative tests to evaluate productivity and systematicity in Neural Machine Translation (NMT). For both productivity and systematicity, we first show that a standard Seq2Seq model performs poorly under the tests for evaluating productivity and systematicity. Further, we investigate the encoder representations, which constitute the first potential bottleneck for compositionality in the sequence to sequence transduction pipeline as a potential cause for poor performance on these tests. We quantitatively demonstrate the weaknesses of encoder representations and propose a simple pre-training scheme to improve the performance on the two properties, leading to considerable improvements in BLEU scores.

\section{Evaluating Productivity and Systematicity in NMT}
 For evaluating productivity of a neural model, \citet{hupkes2019compositionality} propose to evaluate the model on longer sentences than what is observed during training. However, we argue that the concept of productivity is not adequately captured by only evaluating the model against longer sentences. Specifically, we distinguish between the closely related problem of modeling long term dependencies \citep{vaswani2017attention}, versus "unbounded" recombination of already learnt or known rules. Therefore, as an actual test of productivity, we relax the generation of longer outputs to the case when the input is a concatenation of two sentences, on which the model is already known to perform well individually. This relaxation step enforces the model to produce output for similar content, the likes of which the model has already seen within the current temporal processing  step. This evaluation condition, therefore, requires the model to discriminate between short and long term contexts, which is required to identify rules which could be applied repeatedly, within a given context. Further, to evaluate systematic generalizability or systematicity in NMT, \citet{lake2017still} proposed a simple test, wherein the test set contains context for which the model has not seen a particular word. We adopt the same test for evaluating systematicity as well.

% \begin{itemize}
%     \item \textbf{Bag-of-Word Test} (BOWT)- The encoder hidden states of a trained seq2seq network are used to predict a bag-of-words representation of the input sequence. Precision, Recall and F1 are the metrics for evaluating the performance.
%     \item \textbf{Substring Test} (SUBST) - Given two encoder hidden states $h^x$ and $h^y$ which is generated from sentences $x$ and $y$ respectively, the task is to predict whether sentence $x$ is a substring of sentence $y$. The input to this model is $[h^x;h^y]$.
%     \item \textbf{Internal Observation Test} (IOBT): This test measures the similarity between encoder hidden representation captured at the final two contiguous timesteps.
% \end{itemize}

\section{Experiments and Analysis}
\label{analysis}

In this section, we quantitatively demonstrate inadequate temporal processing in the form of poor encoder representations as a bottleneck for compositionality in the sequence to sequence transduction pipeline. All experiments are done using a standard Seq2Seq model with one layer LSTM as encoder and decoder, along with the general attention mechanism \citep{luong2015effective}.

\subsection{Productivity}
\label{bowww}
\textbf{Evaluation:} We train a seq2seq model with attention on the IWSLT-14 German-English dataset \citep{cettolo2014report}, observing a BLEU score of 30.38. We then sample $100$ translations from the validation set, on which we observe a BLEU score of 30.19. We construct a dataset of long sequences by concatenating \textit{pairs of sentences} from the aforementioned sampled data. Evaluating the seq2seq model on this dataset leads to a BLEU score of 27.50 which is a drop of \textbf{2.69 BLEU} points. Though capable of producing quality translations of single sentences, the model performs much worse when tasked with translating two sentences simultaneously, despite the inherent difficulty of the task remaining unchanged.

Based on the above mentioned observation and also to further analyze the encoder representations, we perform two experiments on this dataset of concatenated sentences. For both these experiments we utilise a 3 layer feed forward network with \textit{tanh} activations in the intermediate layers.

\textbf{Experiment 1:} For the first experiment, we use the encoder hidden representations to predict the bag-of-words representation of the input sequence. Specifically, given a sentence $w_{1\cdots t}$ we use the Seq2Seq network to obtain a hidden state $h_t$ in order to produce a bag-of-words representation $b_t$, and evaluate the quality of our hidden representations by attempting to produce $b_t$ conditioned on $h_t$. The results for this experiment is provided in Table \ref{bow}. The first row in Table \ref{bow} evaluates the ability of the trained feed-forward neural network to reproduce $b_t$ conditioned on $h_t$. The second and third row evaluate the ability of the model to reproduce $b_i$ (the bag-of-words representation of the first sentence $x$) conditioned on $h_t$ and $h_i$, respectively. We find that despite being able to reproduce $b_i$ at $h_i$, the encoder forgets information by the final hidden-state $h_t$.

\textbf{Experiment 2:} Next, we design an experiment where the feed-forward network is given two hidden states $h^x$ and $h^y$, generated from two concatenated sentences $x$ and $y$ respectively, and is tasked with predicting whether sentence $x$ is a substring of sentence $[x;y]$ ($x$ concatenated with $y$). The input to the model is $[h^x;h^y]$.  The results for this experiment are provided in Table \ref{substring}.  The results depict that we can more accurately determine that the second sentence is a substring of the concatenated sentence, which in turn suggests that the representation $h_t$ better encodes the final elements in the sequence rather than the initial ones.

Both of these experiments demonstrate that the encoder hidden states are a `forgetful' representation of the input sequence.

\begin{table*}
\begin{center} 
\small
    \begin{tabular}{ | c | c | c | c  |}
    
    \hline
    \textbf{Experiment} & \textbf{Precision} & \textbf{Recall} & \textbf{F1} \\ \hline
    Full Sentences & 27.46 & 41.02 & 32.72 \\ 
    Final Hidden State & \textbf{22.46} & \textbf{33.31} & \textbf{26.83} \\ 
    Intermediate Hidden State & 27.53 & 42.39 & 33.38 \\ \hline
    \end{tabular}
      %\vspace{0.5em}
      \caption{The results for the bag-of-words experiment (Experiment 3.1.1).}
      \label{bow}
\end{center}
\end{table*}

\begin{table}
\begin{center}    
    \small
    %\vspace{0.3em}
    \begin{tabular}{ | c | c | c | c |}
    \hline
    \textbf{Experiment} & \textbf{Result} \\ \hline
    Overall Classification Accuracy & 68.94 \\ 
    First Sentence Recall & \textbf{55.55} \\ 
    Second Sentence Recall & 65.65 \\ \hline
    \end{tabular}
    %\vspace{0.5em}
    \caption{The results for the substring experiment (Experiment 3.1.2).}
    \label{substring}
    %\vspace{-2.0em}
\end{center}
\end{table}

\subsection{Systematicity}
\label{systematic}
\textbf{Evaluation:} \cite{lake2017still} demonstrated that Seq2Seq networks struggle with zero-shot generalization, particularly with respect to \textit{systematic compositionality}. They first trained a translation model on a simple English-French (en-fr) dataset \citep{lake2017still,bastings2018jump}, then supplemented the dataset with many repetitions of a simple sentence containing a new word ("i am \textit{daxy}", "je suis \textit{daxiste"}) and continued training. The original dataset is very simple, with all the sentences beginning with English phrases such as “I am,” “he is,” “they are,” and their contractions. 

They observed that despite seeing occurrences of the word \textit{daxy}, at test time the model could not generalize to constructions such as \textit{"you are daxy"} or \textit{"he is very daxy"}, as it had only seen \textit{"daxy"} in a single context. This lack of systematic compositionality precludes translation models from effectively learning on small amounts of data and as such, an improvement to compositionality could have a potential impact in low-resource machine translation. 

\begin{table}
        \centering
        \begin{tabular}{ | c | c | c |}
        \hline
        \textbf{Word (In template)} & \textbf{Avg. Similarity}  \\ \hline
        daxy & 0.949 \\
        tall &  0.855 \\ 
        ok &  0.887\\ 
        fat &  0.882 \\ 
        fit & 0.878\\ \hline
        \end{tabular}
        %\vspace{0.3em}
        \caption{Results for the average cosine similarity of the encoder representation before and after observing the word in the template (Experiment 3.2.1). It is clear that the average similarity for the word `daxy' is quite higher than the other words. This implies that even after observing the word `daxy', the hidden representation of the encoder did not change much.}
        \label{tab:end_word}
    \end{table}

\begin{table}
        \centering
        \begin{tabular}{ | c | c | c | c  |}
        \hline
        \textbf{Experiment} & \textbf{Precision} & \textbf{Recall} \\ \hline
        Previously \textit{seen} sentences & 0.7 & 0.68 \\ 
        Previously \textit{unseen} contexts & \textbf{0.4} & \textbf{0.4}  \\ \hline
        \end{tabular}
        %\vspace{0.3em}
        \caption{We evaluate the bag-of-word model's ability to produce a bag-of-words representations which contains the word ``daxy,'' both on sentences that were \textit{seen} and \textit{unseen} during training. We observe that the representative power of the encoder hidden states is significantly stronger for sentences the encoder has seen than for unseen sentences, indicating that the encoder hidden representations don't effectively generalize.}
        \label{tab:bow_daxy}
    \end{table}
    
% In order to further test the hypothesis of weak encoder representations, we perform a set of three experiments.
\textbf{Experiment 1:} For the first experiment we constructed a template of six sentences: ``you are $X$", `he is very $X$", ``i am very $X$", ``he is not $X$", ``i am not $X$" where $X$ $\in$ \{daxy, tall, ok, fat, fit\}. For a given $X$ and a template, we measured the cosine similarity of the encoder representation before and after observing $X$. The results for this can be seen in Table \ref{tab:end_word}. For each word, the similarity score was aggregated over all template instances. We observe that the average similarity for the word ``daxy" is much higher than the others, suggesting that the encoder hidden state remains largely unchanged even after observing ``daxy.''

\textbf{Experiment 2:} For the second experiment, the task was to predict the penultimate word of a sentence conditioned on the final encoder representation. Training set for this is constructed using two sentence templates: ``i am $X$" and ``you are $X$", where $X$ is chosen from a set of 128 words, including ``daxy". The test set is the same as in the first experiment above and contains different sentence templates. Given a representation of a sentence $X$, $h^X$, we train a linear layer to produce a probability distribution over the set of all possible values of $X$. Our experiment shows that for ``daxy", the accuracy of the model was 16.6\%, while for all other words it was 50\%. Therefore, the results for both these experiments demonstrate that the encoder hidden state fails to capture the occurrence of ``daxy'' in the input, thereby precluding successful translation.

\textbf{Experiment 3:} For the third experiment, we employ a similar bag-of-words experiment as described in Section \ref{bowww}. We use the last hidden state of the encoder and train a model to predict a bag-of-words representation of the input sequence. We evaluate the ability of the model to produce a bag-of-words representations which contain the word ``daxy,'' both on sentences that were \textit{seen} and \textit{unseen} during training. The results, shown in Table \ref{tab:bow_daxy}, demonstrate that both the precision and recall is significantly lower in \textit{unseen} (by the encoder) sentences, which indicates that the encoder representation does not effectively generalize and is much weaker on unseen sentences.
% The results of all these three experiments provide more substance to our initial hypothesis of weak encoder representations.
% % \textbf{1.} 

\section{Proposed Solution}
Multiple experiments in Section \ref{analysis} strongly validate our hypothesis that the encoder representations are unable to effectively model the input. In order to mitigate this issue we propose a simple \textbf{bag-of-words pre-training} mechanism.

\begin{table*}[!htbp]
\begin{subtable}[t]{0.5\textwidth}
\small
\centering
 \begin{tabular}{ | c | c | c | c |}
    \hline
    \textbf{Method} & \textbf{BLEU} \\ \hline
    Baseline Sequence-to-Sequence &  24.5\\ 
    \textbf{Bag-of-Words Pre-training} & \textbf{26.4} \\ \hline
    \end{tabular}
    %\vspace{0.3em}
      \subcaption{}
    \label{tab:scores}
%     %\vspace{-1.5em}
\end{subtable}
\begin{subtable}[t]{0.5\textwidth}
\small
 \begin{tabular}{ | c | c | c | c |}
    \hline
    \textbf{Method} & \textbf{BLEU} \\ \hline
    Baseline Sequence-to-Sequence & 36.14 \\ 
    \textbf{Bag-of-Words Pre-training} & \textbf{52.53} \\ \hline
    \end{tabular}
    %\vspace{0.3em}
      \subcaption{}
    \label{tab:daxy_dual_bow}
\end{subtable}
\caption{\textbf{\ref{tab:scores})} : Comparison of our proposed method over the task of translating sentence pairs (Productivity). These results are statistically significant with $p = 0.011$. \textbf{\ref{tab:daxy_dual_bow})} : The performance of our proposed methods on the task of systematicity. We notice that bag-of-words pre-training gives a significant improvement, suggesting that better modelling the input is an effective mechanism for strengthening the encoder representations.}
\label{final}
\end{table*}

In the proposed approach of bag-of-words pre-training, we first train our encoder to encode a sentence $S$, to produce $h^s$, and predict the bag-of-words representation of $S$ conditioned on $h^s$. More specifically, we use $h^s$ as input to a linear layer with $V$ hidden units to produce $X$, where $V$ is the number of terms in the vocabulary. Following this, the loss can be computed as follows:
\begin{equation}
\small
      \ell = -\bigg(\sum_{i = 1}^{V}y_n\log\sigma(x_n)  + (1 - y_n\log\sigma(1 - x_n))\bigg)
\end{equation}
where $\sigma$ is the sigmoid function, $x_i$ is the $i^{th}$ element of $X$ and $y_i$ indicates whether the $i^{th}$ word is present in the sentence $S$ or not.

Given this pre-trained encoder, we then initialize our Seq2Seq network for machine translation. Given that our analysis demonstrated the weakness of the encoder representations, this pre-training should strengthen the representations by initializing them in a way that forces it to effectively capture the input sentence. The results for this can be seen in Table \ref{final}. On both the compositionality tasks, our proposed solution leads to siginificantly superior performance. Further, on the task of systematicity, we observe that using the pre-trained encoder, the NMT model is able to generate the word `daxy' (in the translated sentence) in \textbf{$50\%$} of cases, which is considerably higher than the baseline at \textbf{$12.25\%$}.

\vspace{-0.4cm}
\section{Conclusion and Future Work}
We demonstrated that poor temporal processing in the form of inadequate encoder representations in Seq2Seq models do not lead to a good performance on tasks requiring compositionality. We presented a simple, though well-motivated, approach for improving the encoder representations. A potential avenue for future work may be to explore more sophisticated mechanisms of improving the representative power of encoder hidden states, including larger-scale pre-training, multi-task learning, and more complex pre-training objectives. More generally, our analysis demonstrates that, through carefully designed experiments, the failures of Seq2Seq models can be attributed to specific components of their internal mechanisms, thereby allowing for more targeted and better motivated architectural improvements. Even though our experimentation was focused on enhancing compositionality, our analytic methods could be applied to analyze other failures of Seq2Seq models as well. Both from the perspective of model analysis and architectural improvement, this work makes significant progress and opens up several avenues for future research.

\bibliography{bib}

\end{document}